\documentclass[10pt,twocolumn,letterpaper]{article}

\usepackage[pagenumbers]{cvpr} 

\usepackage{booktabs}

\usepackage[T1]{fontenc}

\usepackage{microtype}

\definecolor{cvprblue}{rgb}{0.21,0.49,0.74}
\usepackage[pagebackref,breaklinks,colorlinks,allcolors=cvprblue]{hyperref}

\def\paperID{*****} 
\def\confName{CVPR}
\def\confYear{2026}

\title{CuriosAI Submission to the CASTLE Challenge at EgoVis 2026}

\author{
Yuto Kanda\thanks{Equal contribution.}
\and
Hayato Tanoue\footnotemark[1]
\and
Takayuki Hori\\
SoftBank Corp.\\
{\tt\small
\{yuto.kanda, hayato.tanoue, takayuki.hori\}@g.softbank.co.jp
}
}

\begin{document}
\maketitle
\begin{abstract}
CASTLE 2026 asks $185$ multiple-choice questions over $600+$ hours of
synchronised multi-view egocentric video. We explore two approaches on
top of a shared multimodal preprocessing layer (per-person timelines,
speaker-resolved transcripts, multi-VLM caption ensembles, etc.).
\textbf{Approach A (SVA: Search--Verify--Answer)} is a three-stage
pipeline that hierarchically narrows to a primary window, verifies
sub-windows with a VLM under four anti-confabulation rules, and fuses
evidence with an LLM judge under an evidence-priority hierarchy.
\textbf{Approach B (TMKG: Temporal--Multimodal--Knowledge--Graph)} is
the contrast: it builds a temporal multimodal knowledge graph, locates
a primary cell via graph search, and produces the final answer with a
single grounded VLM. SVA reaches a leaderboard accuracy of $0.50$ and
is our final challenge submission; TMKG reaches $0.35$.
\end{abstract}

\section{Introduction}
\label{sec:intro}

In this paper, we present our submission to the CASTLE Challenge at
EgoVis 2026. The challenge asks $185$ multiple-choice questions over
$600+$ hours of synchronised multi-view egocentric (\emph{ego}) and
exocentric (\emph{exo}) footage from the CASTLE 2024
dataset~\cite{rossetto2025castle}, coupling \emph{long-form retrieval}
with \emph{fine-grained multimodal verification} --- infeasible for a
single VLM at this scale.

A second obstacle is that an unconstrained video-audio specialist
frequently \emph{confabulates}: it re-quotes prompt context, picks an
option on silent or test-pattern clips, and asserts counts without
spatial grounding. These failure modes motivate the prompt-level
discipline central to our design.

Building on these observations, we pursue two complementary directions
on top of a shared preprocessing layer. \textbf{SVA}
(\S\ref{subsec:approachA}) frames retrieval as cell-indexed search and
isolates the final judgement behind an explicit anti-confabulation
discipline. \textbf{TMKG} (\S\ref{subsec:approachB}) is the contrast:
it tests whether knowledge-graph retrieval paired with a single
grounded VLM consuming the selected cell's multi-camera video frames
and bundled evidence (captions, transcripts, tags) can replace
cell-indexed search and the decoupled verify--judge stack. On the
leaderboard, SVA $0.50$ vs TMKG $0.35$ indicates that verifier
discipline contributes more than altering the retrieval structure at
the CASTLE 2026 scale. SVA is our final submission.

\section{Task}
\label{sec:task}

The footage covers four days of shared-living recordings from $12$
participants, with $10$ head-mounted ego cameras (worn per day) and
five fixed exo cameras yielding $15$ synchronised $4$K video streams
with aligned audio. The $185$ questions are four-choice; the official
metric is the accuracy averaged over all questions.

\section{Method}
\label{sec:method}

Our two approaches share a common multimodal preprocessing layer
(\S\ref{subsec:preprocess}). On top of it, SVA decouples retrieval,
verification, and judgement into three explicit stages, while TMKG
packages observations into a temporal knowledge graph and produces the
final answer from a single grounded VLM
(Fig.~\ref{fig:approaches}).

\subsection{Shared Preprocessed Multimodal Databases}
\label{subsec:preprocess}

\begin{figure*}[t]
\centering
\includegraphics[width=\linewidth]{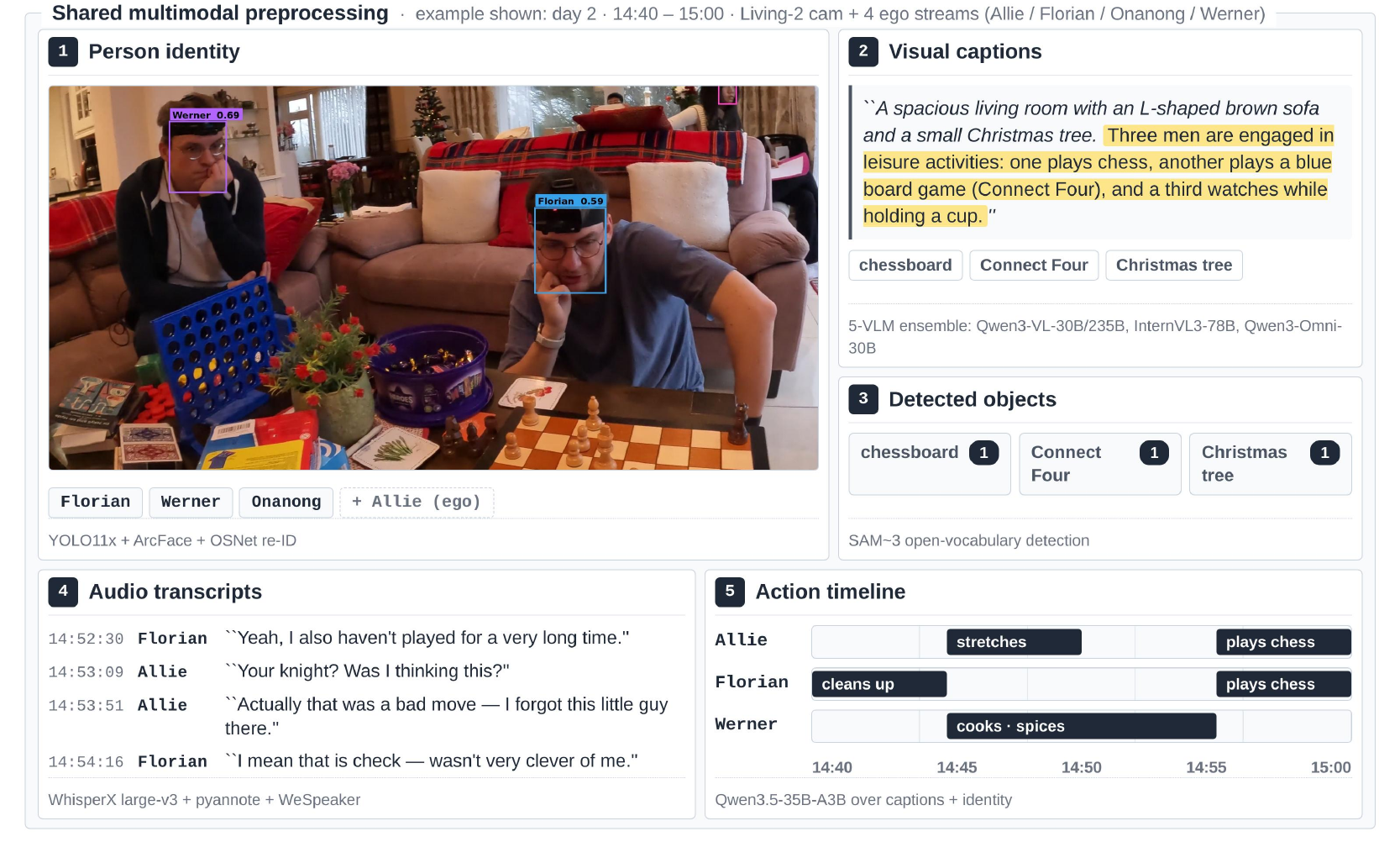}
\caption{%
Shared preprocessing substrate. Five offline lanes over the CASTLE 2024 corpus ($15$ cameras, $\approx$$600$\,h of $4$K video and audio) aligned on a common temporal axis. Both approaches consume this layer read-only; their approach-specific aggregations are described in \S\ref{subsec:approachA} (SVA) and \S\ref{subsec:approachB} (TMKG).
}
\label{fig:preprocess}
\end{figure*}

Five preprocessing lanes are built offline from the publicly released
CASTLE 2024 footage and consumed read-only by both approaches through
approach-specific aggregation (Fig.~\ref{fig:preprocess}). We describe the lanes in the order shown
in Fig.~\ref{fig:preprocess}: person identity, visual captions,
detected objects, audio transcripts, and action timeline.

\paragraph{Person identity.}
YOLO11x detects bodies; ArcFace~\cite{deng2019arcface} face embeddings are matched against twelve participant centroids built from frontal reference images, with cross-camera propagation gated by OSNet~\cite{zhou2019omni} body re-identification requiring body--body similarity and face--centroid agreement.

\paragraph{Visual captions.}
We run four captioner models in five settings: Qwen3-VL-30B-A3B-Instruct at $300$\,s windows (scene) and $1800$\,s windows (temporal narrative); InternVL3-78B~\cite{zhu2025internvl3} with a verb-focused prompt for action descriptions; Qwen3-Omni-30B~\cite{xu2025qwen3omni} for joint video+audio captioning; and Qwen3-VL-235B-A22B-Thinking-AWQ~\cite{bai2025qwen3vl} for reasoning-heavy captions.

\paragraph{Detected objects.}
An open-vocabulary detection lane is produced by SAM~3~\cite{carion2025sam3}. CASTLE 2026 questions frequently reference objects outside generic ImageNet/COCO vocabularies (e.g., specific board games, kitchen equipment, holiday decorations), so open-vocabulary detection is required.

\paragraph{Audio transcripts.}
WhisperX~\cite{bain2023whisperx} (large-v3~\cite{radford2023robust}) transcribes, pyannote~\cite{plaquet2023powerset} diarises, and WeSpeaker~\cite{wang2023wespeaker} resolves speakers. On fixed exo cameras a \emph{fixed-camera consensus} rule restricts speaker candidates to those corroborated by both a concurrent transcript on another camera and the identity lane.

\paragraph{Action timeline.}
A per-person \emph{action timeline} is summarised by Qwen3.5-35B-A3B~\cite{qwen2026qwen35} from the captions and identity lane as (time-span, verb, co-actors) tuples. These five lanes form the shared layer; their approach-specific aggregation is described in \S\ref{subsec:approachA} (SVA) and \S\ref{subsec:approachB} (TMKG).

\subsection{Approach A: SVA (Search--Verify--Answer)}
\label{subsec:approachA}

\begin{figure*}[t]
\centering
\includegraphics[width=\linewidth]{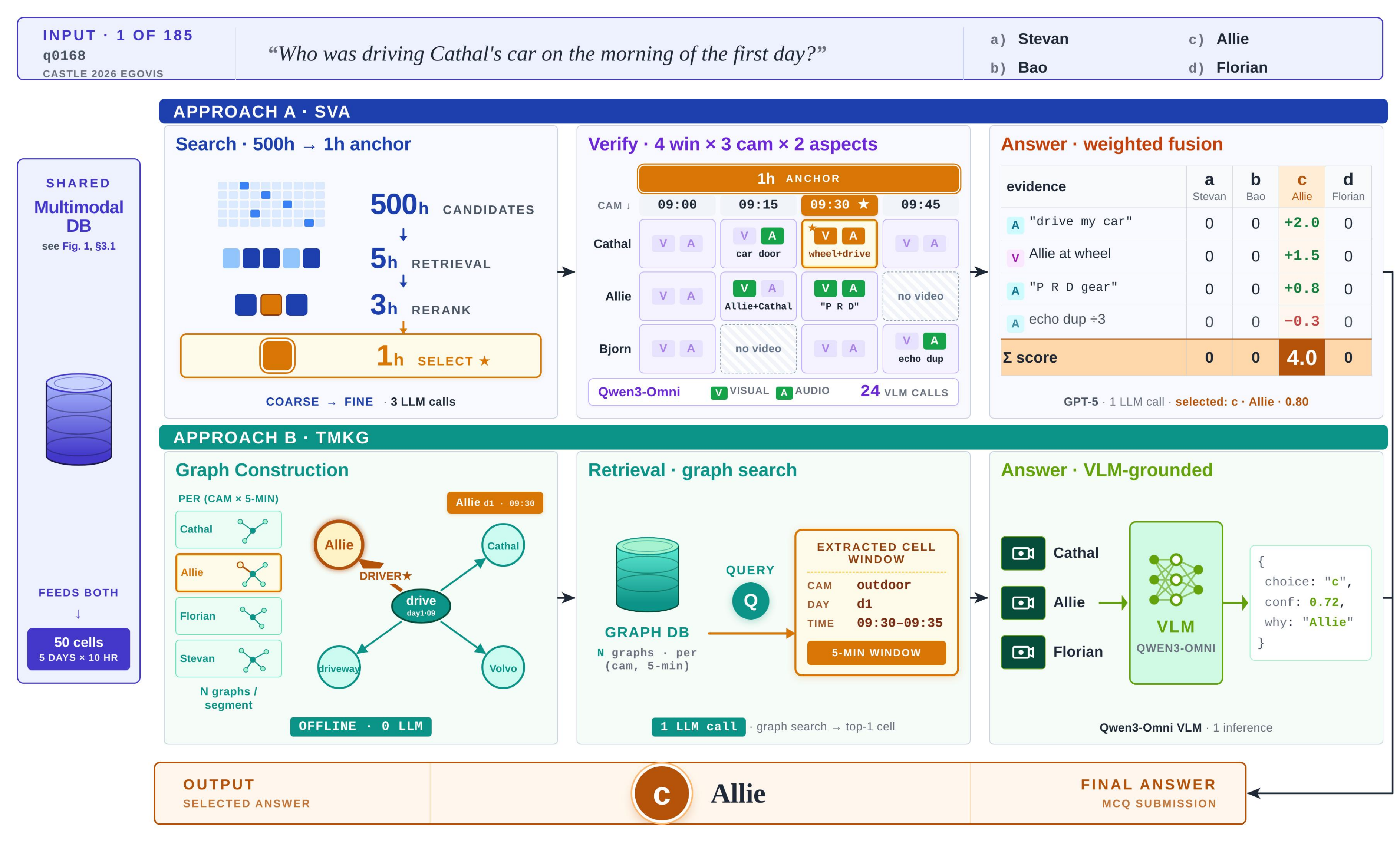}
\caption{%
Walk-through of both pipelines on a sample question (q0168), reading the shared DB (Fig.~\ref{fig:preprocess}). \textbf{SVA} (top): narrow $\sim$$50$ cells, verify $\approx 24$ sub-windows under anti-confabulation rules, fuse with a GPT-5 judge. \textbf{TMKG} (bottom): build and search a temporal knowledge graph, answer with a single grounded Omni VLM.
}
\label{fig:approaches}
\end{figure*}

SVA is a three-stage pipeline (Fig.~\ref{fig:approaches}, top row):
\textbf{Search} narrows the candidate cells of \S\ref{subsec:preprocess}
to one primary $15$-minute window; \textbf{Verify} extracts evidence
under anti-confabulation rules; \textbf{Answer} fuses it with a single
LLM judge.

\paragraph{Search.}
SVA aggregates preprocessing into $50$ \emph{(day, hour)} cells,
summarises each with Qwen3.5-35B-A3B, and indexes via
BM25~\cite{robertson2009probabilistic} +
e5-large-v2~\cite{wang2022text} hybrid retrieval. A
GPT-5-mini~\cite{openai2025gpt5} reranker and $15$-min bucket scorer
narrow the candidates; a final GPT-5 reasoning call picks the primary
window, cross-camera supporting windows, and a tentative answer carried
into Answer as a prior.

\paragraph{Verify.}
The Search anchor is expanded into $\approx$$24$ adjacent $5$-minute
sub-windows across cameras, each verified by
Qwen3-Omni-30B-A3B-Instruct. Unconstrained, the
verifier exhibits four characteristic confabulations (re-quoting prompt
context, asserting choices on silent or test-pattern clips, counting
without spatial grounding, ungrounded high confidence), which we
suppress with four \emph{anti-confabulation rules} in the system
prompt: \emph{no echo}, \emph{abstain}, \emph{localise}, and
\emph{ground}.

\paragraph{Answer.}
A single GPT-5 judge fuses the question, choices, ranked per-window
evidence, the Search-stage tentative answer, and a
Gemini-2.5-Pro~\cite{comanici2025gemini} reasoning trace on the
narrowed clip as an external prior. The judge follows an explicit
evidence-priority hierarchy
(OCR\,$\succ$\,audio quote\,$\succ$\,visual\,$\succ$\,context),
deduplicates echoed quotes, and rejects counting answers that lack
spatial localisation.

\subsection{Approach B: TMKG (Temporal--Multimodal--Knowledge--Graph)}
\label{subsec:approachB}
TMKG (Fig.~\ref{fig:approaches}, bottom row) consumes the preprocessing
lanes (\S\ref{subsec:preprocess}) at a $5$-minute per-camera granularity
and operates in three layers: \textbf{Graph Construction} builds and
persists a temporal knowledge graph; \textbf{Retrieval} narrows the
evidence cells via hybrid indexes and graph constraints; and
\textbf{Answer} produces the final choice with a single grounded VLM.

\paragraph{Graph Construction.}
For each \emph{(camera, $5$-minute cell)} we bundle the concurrent
transcript, captions, actions, objects, and summary tags into a single
\textbf{Observation} node, and aggregate multi-view Observations that
share time, location, people, actions, and visual features into a
single \textbf{GlobalEvent}. Typed edges link
Observation\,$\to$\,Event (evidence), Person\,$\to$\,Event
(participation), Event\,$\to$\,Place/Object (location/involvement), and
consecutive Events to each other (\texttt{PRECEDES}). Each
\emph{(date, time)} segment is persisted independently.

\paragraph{Retrieval.}
A query parser extracts date, time, person, place, object, action, and
intent from the question and choices.
BM25 and e5-large-v2 are run over Qwen3-VL-30B-A3B-Instruct summaries of the
Observations and fused via reciprocal-rank fusion~\cite{cormack2009reciprocal}. Graph
predicates (\texttt{PARTICIPATES\_IN}, \texttt{LOCATED\_AT},
\texttt{INVOLVES}) rerank relational queries, and the
\texttt{PRECEDES} edge from the top cell handles ``immediately
before/after'' questions. If the top score and margin clear thresholds,
a single cell is passed to the answer layer; otherwise a concatenation
of the top cells is passed.

\paragraph{Answer.}
The selected cell's multi-camera video frames and bundled multi-camera
evidence (captions, transcripts, tags) are passed with the question and
choices to a single Qwen3-Omni-30B-A3B-Instruct, which emits the
choice, confidence, supporting camera views, and rationale as JSON. A
three-stage fallback (alternative VLM\,$\to$\,caption excerpts\,$\to$\,
default value) covers primary-model failures.









\section{Results and Conclusion}
\label{sec:conclusion}

\paragraph{Results.}
Table~\ref{tab:lb} reports the official leaderboard accuracy: SVA
reaches $0.50$ (our final entry); TMKG reaches $0.35$. SVA issues
roughly $28$ LLM/VLM calls per question vs roughly $1$ for TMKG.
Across both pipelines, the dominant failure mode is the Search stage
narrowing to a wrong cell that downstream verification cannot recover
from. The anti-confabulation rules are not ablated; they should be read
as a design lesson rather than an isolated measured gain.

\begin{table}[t]
\centering
\small
\begin{tabular}{lc}
\toprule
Method & LB \\
\midrule
\textbf{Approach A (SVA, ours)} & \textbf{0.50} \\
Approach B (TMKG) & 0.35 \\
\bottomrule
\end{tabular}
\caption{CASTLE 2026 official leaderboard accuracy on the $185$-Q evaluation set. Approach~A is our final challenge entry.}
\label{tab:lb}
\end{table}

\paragraph{Conclusion.}
At CASTLE 2026 scale, the $0.15$ SVA--TMKG gap suggests that
disciplined clip-level verification (abstention, count localisation,
evidence grounding, no-echo) is a layer worth adding on top of any
retrieval substrate, at the cost of noticeably more LLM/VLM calls.
Since both pipelines are bounded by the same Search-stage
cell-localisation failures, combining higher-precision retrieval with
this discipline is the natural next direction.

{
    \small
    \bibliographystyle{ieeenat_fullname}
    \bibliography{main}
}


\end{document}